# Modified Fast Fractal Image Compression Algorithm in spatial domain

M.Salarian, H. Miar Naimi
Electrical Engineering Faculty
University Of Alame Mohades Noori
Tel-Fax: 98-122-6253521
***Email: mehdi.salarian@gmail.com, h_miare@nit.ac.ir***

## Abstract

In this paper a new fractal image compression algorithm is proposed in which the time of encoding process is considerably reduced. The algorithm exploits a domain pool reduction approach, along with using innovative predefined values for contrast scaling factor, ***S***, instead of searching it across [0,1]. Only the domain blocks with entropy greater than a threshold are considered as domain pool. As a novel point, it is assumed that in each step of the encoding process, the domain block with small enough distance shall be found only for the range blocks with low activity (equivalently low entropy). This novel point is used to find reasonable estimations of ***S,*** and use them in the encoding process as predefined values, mentioned above, the remaining range blocks are split into four new smaller range blocks and the algorithm must be iterated for them, considered as the other step of encoding process. The algorithm has been examined for some of the well-known images and the results have been compared with the state-of-the-art algorithms. The experiments show that our proposed algorithm has considerably lower encoding time than the other where the encoded images are approximately the same in quality.

## 1. Introduction

Fractal image compression is widely used in image processing applications such as image signature [1], texture segmentation [2,3], feature extraction [4], image retrievals [5,6] and MR, ECG image processing [7,8,9]. The interesting advantages of the fractal compression are fast image reconstruction and high compression ratio. Another advantage of fractal image compression is its multi-resolution property, i.e. an image can be decoded at higher or lower resolutions than the original, and it is possible to “zoom-in” on sections of the image [10]. These properties made it very suitable for multimedia applications, so that it was used by Microsoft to compress thousands of images in its multimedia encyclopedia [7]. In spite of all above advantages, fractal image coding suffers from long encoding time that still is its main drawback. This long encoding time arise from very large number of domain blocks that must be examined to match each range block. The number of range blocks with size of $n \times n$, in an $N \times N$ image, is $(N/n)^2$, while the number of domain blocks is $(N-2n+1)^2$. Consequently it can easily be shown that the computation for matching range blocks and domain blocks has complexity of $O(N^4)$ [12]. Thus focus of researches is to reduce the encoding time. Several methods have been proposed to overcome this problem. One of the common ways is the classification of blocks in a number of distinct sets where range and domain blocks of the same set are selected for matching, Here the encoding time is saved at cost of losing the quality. The Fisher classification method can be addressed as a good example [1,10,11].

Reducing the size of domain pool is another method that has been done in different ways, in some researches domain blocks with small variance [4] and in some others domain blocks with small entropy were deleted from domain pool [12]. Another approach covers the hybrid methods that use spatial domain and frequency domain information to compress images [19,20]. In addition to the size of domain pool, the computational cost of matching a range block and a domain block has an important role in encoding time. We reduced this cost by estimating the approximate optimum values for contrast scaling factor, ***S,*** instead of searching for it**.** Combining these two novel points, we propose a new fractal image coding that have considerable shorter encoding time than the last fast algorithm [12]. The rest of the paper is as follows, section 2 introduce a brief description of the fractal image coding. The proposed algorithm is presented in section 3. In section 4 the experiments and the results are presented and compared with the last algorithm. Finally in section 5 conclusions are presented and some future works are addressed.

## 2. Fractal Image Coding: A Brief Review

At the first step in fractal coding the in hand image is partitioned into none overlapping range blocks of size $B \times B$ where, $B$ is a predefined parameter [5,6,13]. Then a set of domain blocks is created from original image, taking all square blocks of size $2B \times 2B$ with integer step L, in horizontal and vertical directions. The minimum value for L is 1 that leaves domain set with maximum size. Related to each member in domain pool, three new domain blocks are created by clockwise rotating it 90º, 180º and 270º, also these three and the original domain block all are mirrored. Here, in addition to the original domain block, we have 7 new domain blocks. To enrich the domain pool and empower the search process, these new 7 domain blocks are added to the domain pool. After constructing the domain pool, related to each range block we must select the best domain block from domain pool and find an affine transformation that maps the selected domain block to it with minimum distance. The distance is taken in the luminance dimensions not the spatial dimension. Such a distance can be defined in various ways but to simplify the computations it is convenient to use Euclidean metric. It must be noted that the distance is taken between range block and the decimated form of domain block, because of larger dimensions of the latter. Each range block is associated to a domain block and the related affine transformation that defines a mapping between them with minimum distance. For each range block the address of related domain block and the affine transformations are stored as the result of compression. In decoding process each range block is constructed from the associated domain block and the transformation. The mentioned distance between a range block, $R$ , and a decimated domain block, $D$ , both with $n$ pixels is defined as follows:

$$E(R,D) = \sum_{i=1}^{n} (Sd_i + O - r_i)^2 \qquad (1)$$

The minimum of the above objective function occurs when the partial derivatives with respect to ***S*** and ***O*** are zero. Solving these two resulting equations will give the best coefficient ***S*** and ***O*** as follows [14]:

$$S = \frac{< R - \overline{R}.1, D - \overline{D}.1 >}{\| D - \overline{D}.1 \|^2} \qquad (2)$$

$$O = \overline{R} - s\overline{D} \qquad (3)$$

$<,>$, *R*, *D*, $\overline{R}$ and $\overline{D}$ are inner product, range block, domain block, mean of *R* and mean of *D* respectively. Because of high computational cost of (2), it is convenient to search ***S*** across a pre-sampled set of [0,1], instead of calculating (2). The above process of encoding can be done totally in one step, ignoring the error values of mappings, leaving fast but very lossy algorithm, indeed the quality and ability of the algorithm is restricted to the existing similarities in the blocks with defined dimension, but higher similarities may be found among smaller regions (equivalently smaller blocks). To have range blocks with changeable sizes a different approach was used in which, the search for finding two similar regions (blocks) are done hierarchically in some steps [12,15]. Along the matching process, the best found transformation only saved for range blocks which, have been mapped with an acceptable error. The remaining range blocks are split into 4 new smaller range blocks, and the matching process is restarted for them as a new step. It is clear that the domain pool of the new step consists of smaller domain blocks. Regarding to the initial size of range blocks, the algorithm may have different number of steps, for example if range blocks initially have size of $16\times16$ pixels, the range blocks of the succeeding steps will have size of $8\times8$, $4\times4$ and $2\times2$ respectively, that leaves a four step algorithm. In each step, a threshold of mapping error determines that the in hand range block must be encoded in current step or split and encoded in the next steps. In a four steps algorithm there are 3 thresholds for the three first steps, the range blocks of the last step are all in size of $2\times2$ that splitting them to smaller range blocks leaves four single pixels that never have any benefit, so at this step the mappings are done anyway and the best transformations found are stored. Figure 1 shows a simple description of all above steps. In fig 1 the range block ABCD was encoded at the first step but range block DEFC was split into four quarter and they all were encoded at step 2, so more data was used to encode the DEFC region. The range block BCKJ could not be encoded at step 1 so it was split into four quarters. Three of four new smaller range blocks of region BCKJ are encoded at step 2 and the last was split into four other quarter or new range block As shown in figure 1, the region BCKJ experienced all 

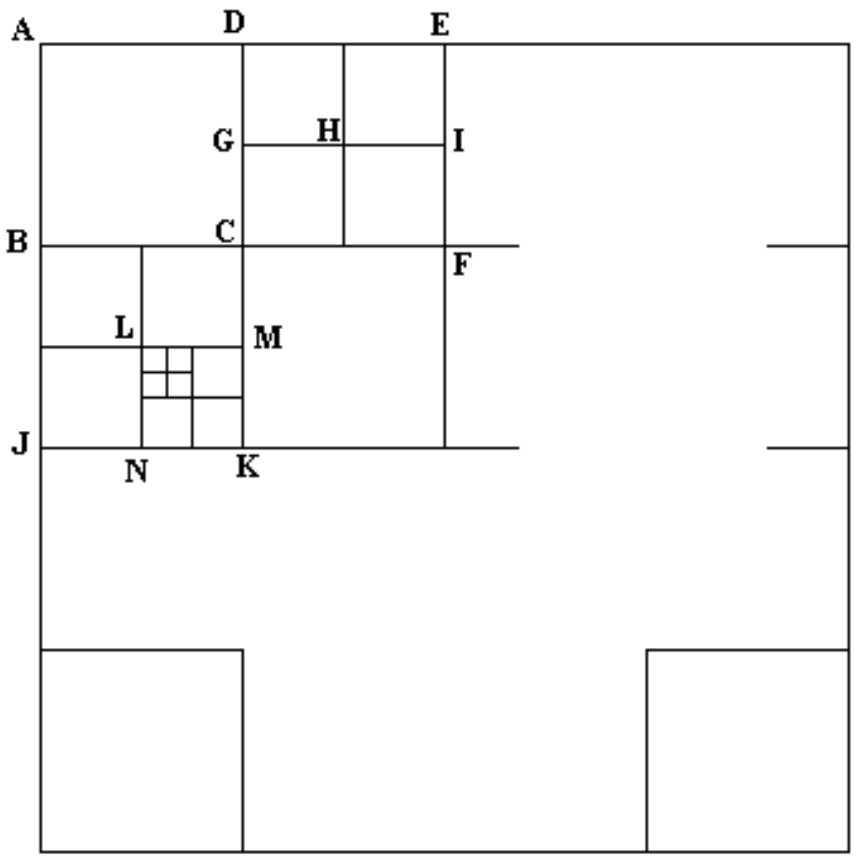

***Fig.1 a simple view of quad tree partitions***

As shown in figure 1, the region BCKJ experienced all 4 steps of the algorithm to be encoded. A good compression algorithm encodes images into very small amount of data in a short time and with minimum losing of information. However there are some tradeoffs among these goodness features, so that enhancing one feature usually necessitates degrading some of the others.To evaluate compression algorithms and also compare every two of them, some measures were defined. The first is the compression ratio that presents how much the coded data is less than the data of original image. The other is PSNR that defines the distance between the decoded image and the original and represents the fidelity of the algorithm. The PSNR is defined as follows:

$$PSNR = 10\log_{10}\left(\frac{255^2}{\frac{1}{N}\sum_{i=1}^{N}(f_i - f_i^*)^2}\right) \qquad (4)$$

Where $N$ is the total number of pixels in image, $f_i$ and $f_i^*$ are $ith$ pixels gray level of the original and decoded images respectively. The other is the encoding time that has very significance in fractal coding, and is defined the total time required to compute compressed data. Fractal image compression algorithms, as mentioned in section 1, have usually long encoding time, so that the current researches focus to reduce it. The last measure is the decoding time that is defined as total time required rebuilding initial image from compressed data, and usually has low values and so little significance in fractal coding. Choosing large values for the mapping error thresholds of different steps causes less range blocks are encoded in the lower steps that saves the encoding time but makes the algorithm lossier. On other words the algorithm will be faster, with lower PSNR and better compression ratio.

Two strategies were used to reduce the encoding time in fractal coding algorithms. In a research Saupe find out that the domain pool is not necessary include all of possible domain blocks and only the high variance blocks are sufficient [4,17]. On the other word he assumed that high variance domain blocks have sufficient ability to be mapped to all range blocks with small enough error, so he deleted all low variance domain blocks from domain pool and used only a small domain pool. In another work like above, the entropy measure was used instead of variance [12, 15]. Entropy and variance have very similar results so we use only the entropy based algorithm as the state the art algorithm. The table below shows the performance of the mentioned methods [16].

***Table 1: comparison between Saupe method and entropy based (DN denotes to the size of domain pool).***

| | bpp (bit per pixel) | Time (Sec) | PSNR(dB) |
|---|---|---|---|
| Entropy(DN=64) | 0.74 | 9.92 | 34 |
| Entropy(DN=256) | 0.67 | 27.2 | 34.8 |
| Saupe(1995) | 0.947 | 39 | 34.57 |

The last row is quoted from [16] and the two other are from our implementations. As shown above the entropy based method is superior to the Saupe method so we compare our algorithm to entropy based one. To understand the significance of the size of domain pool, one can consider an $512\times512$ image, domain block of size $16\times16$ with overlapping of 4 pixels; there are 15625 domain blocks that must be checked to match each range block. The huge number of computations here is obvious. Now, restricting the domain pool to a number of less than 500 members, for example, the above large amount of computation will be decreased about 30 times.

## 3. The proposed algorithm

The proposed algorithm has a global structure like what mentioned in section 2. It has four steps and tries to code the range blocks in the first step with small enough error, otherwise split the range block and continue through other steps as mentioned in previous section. In this paper we use two novel points to reduce the encoding time. The first point is restricting the domain pool to high entropy domain blocks. This causes the total evaluations for finding related domain block of a range block becomes shorter. The entropy of a block is defined as below. Suppose $N$ be a $n\times n$ block of an Image as shown in figure 2.

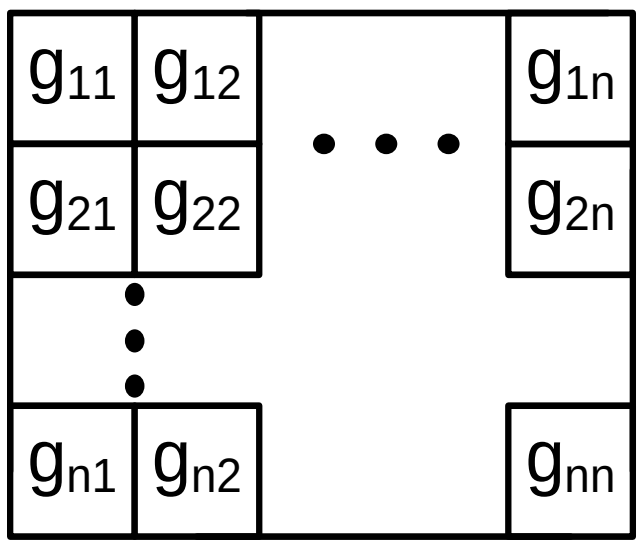

***Figure 2, a domain block of size $n\times n$***

In the above figure $g_{ij}$ is the grey level of the pixel at location $(i, j)$. Suppose $g_{ij}$ for $i, j = 1, 2, ..., n$ varies in $\{L_1, L_2, ..., L_K\}$. Also suppose the number of observations of $L_i$ over the pixels be $q_i$. So the probability of $L_i$ is defined as equation 5,

$$p_i = \frac{q_i}{\sum_{j=1}^{K} q_j} = \frac{q_i}{n^2} \quad (5)$$

The entropy is defined as equation (6):

$$entropy = -\sum_{i=1}^{K} p_i Ln(p_i) \qquad (6)$$

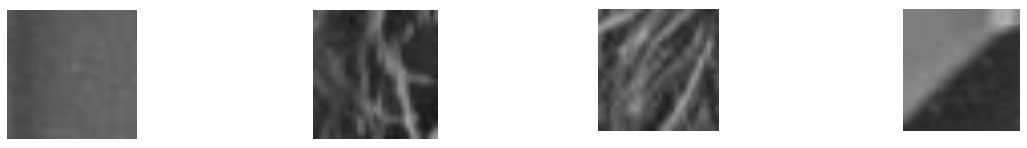

35.7 44.20 46.09 40.57

***Figure 3 four domain blocks and their related entropy***

Shown in figure 3, are some examples of domain blocks and their entropy. As one can find out from figure 3, low entropy blocks are smoother and have less information. Lacking high frequency information, low entropy blocks cannot cover high entropy range blocks. On the other hand high entropy blocks may cover all range blocks. To cover low entropy rang blocks we can simply reduce information of the domain blocks.

## 3.1 The effect of pool size

To have a good insight into the effect of the size of domain pool on PSNR, compression ratio and the encoding time, we did some experiments with different pool sizes, using the entropy based algorithm and similar error thresholds for all experiments. Figures 4a, b, c show the PSNR, compression ratio and the time for Lena respectively, the result for different images have similar pattern like Lena.

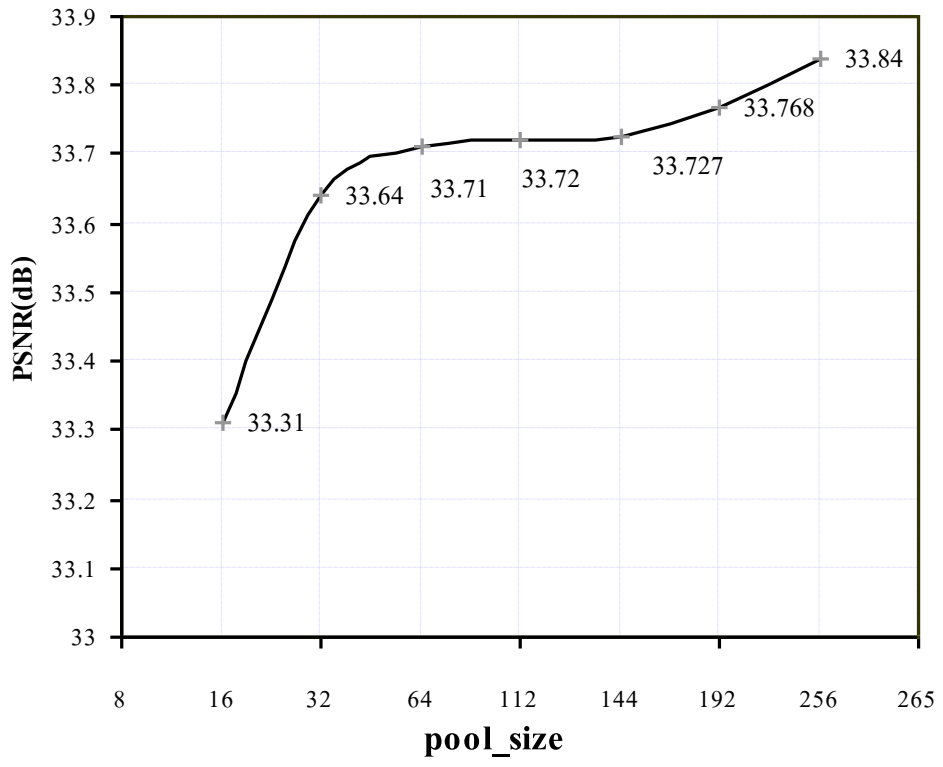

**(4a)**

As one can simply see, when pool size has small values the PSNR and compression ratio are both very small that isn't desirable but grow up very fast by increasing pool size. For greater values of pool size the growth of the two measures are become slow and increasing the pool size only increases the time. These results verify the validity of the domain pool reduction

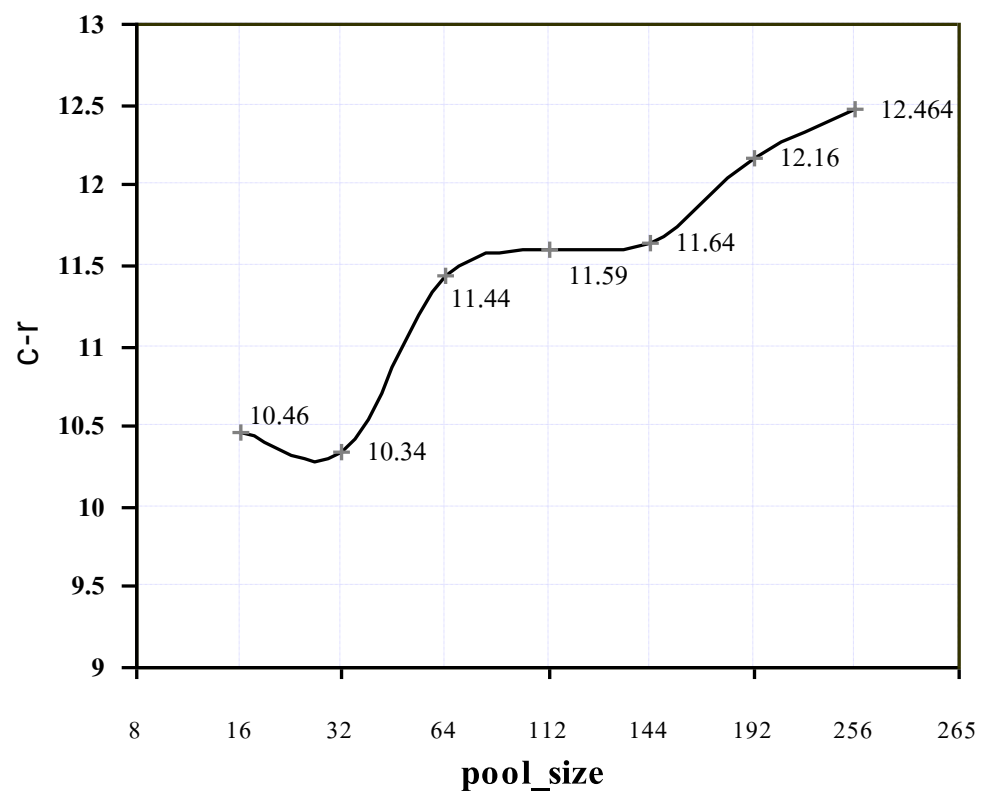

***(4b)***

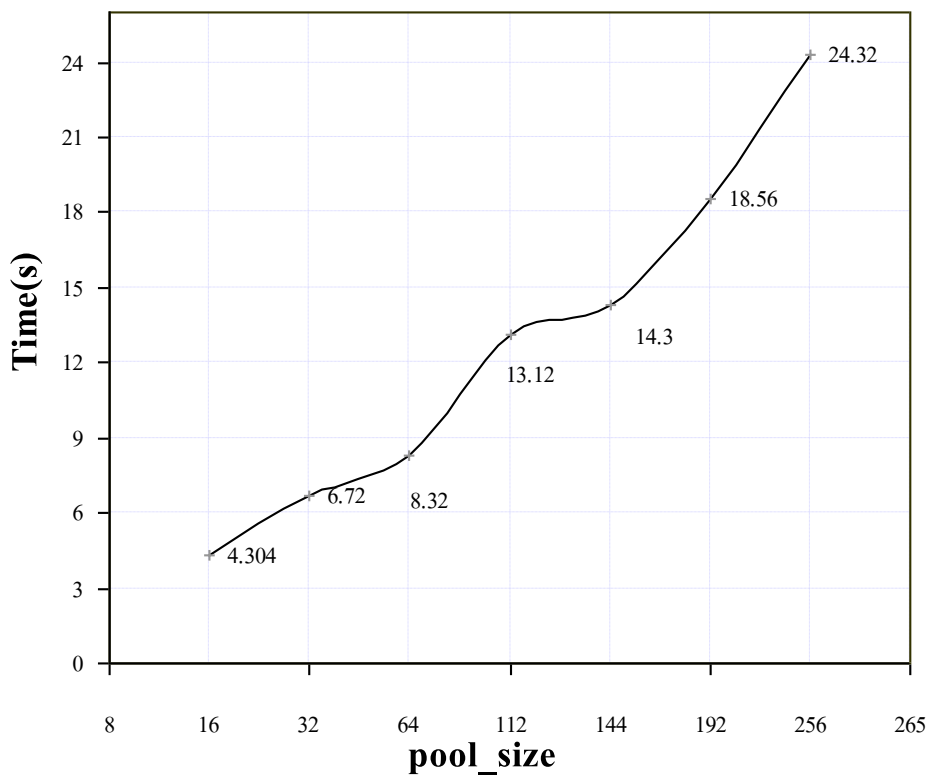

**(4c)**

**Figure 4, a) plot of PSNR versus pool size b) plot of compression ratio versus pool size c) plot of encoding versus pool size**

Approaches. Using these three figures one can approximately tune the performance measures of the algorithm by pool size.

## 3.2 The effect of contrast scaling factor, $s$

Another important parameter that shall be investigated is the contrast scaling factor $s$. To do this a large number of experiments with exhaustive search for $s$ were done. The histograms of the best selected values of $s$ are shown in figure 5 for all four steps respectively. To analysis the effect of $s$ it will be helpful to recall the operation of $s$. As mentioned in section 1 the pixels of domain blocks are multiplied by $s$ and then the integer part is considered. Indeed $s$ maps integer values of domain pixels to integer values of range pixels. At step 1 range blocks are $16 \times 16$ or of size 256 pixels. Consider now a block with a determined entropy or information. It is obvious that all permutation constructed by rearranging pixels of the block has the same entropy as the original. As a simple and qualitative measure or as a lower bound for the number of these permutations is as equation 7.

$$N_P = \frac{256!}{n_1! n_2! \cdots n_k!} \tag{7}$$

Where $n_j$ the number of pixels with grey level of j Big is $n_j$ means the block has simple texture and equivalently low entropy. How much $n_j$ are big the number of distinct permutations is small. As a result of discussion above, at step 1, only range block with small entropy will have the chance to be coded and consequently s has small values (recall proposition above). At step 1 if the entropy of a block is high then, the number of blocks with that entropy will be high ($n_j$ s are small) so with the high probability it cannot be coded at this step. As a result we expect that $s$ has small value. At lower steps 2, 3 and 4 the blocks are in size of $8 \times 8$, $4 \times 4$ where with similar discussion we will see that $N_P$ will drastically decreased as a qualitative comparison we can write:

$$\frac{NP2}{NP1} \propto \frac{64!}{256!} \approx 0$$

So with a similar discussion it can be expected that, blocks of higher entropy at level 2 be encoded that leaves $s$ of greater values. This will also happen in lower steps: the histogram of the best $s$ in lower steps, according above discussion; will be shifted to right, as shown in figure 5. In each step of previous algorithms, all members of a 10-member set of $s$, sampled from [0, 1], are evaluated. It can be seen from figure 5 that all values of $s$ need not be evaluated and we can restrict $s$ to one or two distinct values. It is obvious that restricting the size of the search set of $s$ to a 2-member set will decrease the encoding time considerably. To find true estimation for $s$, a large number of experiments with exhaustive search for S were done. One can easily see that at step 1 the optimal $s$ has often value less than 0.1, independent of the image, so for this step we let $s = 0.1$. At step 2 the optimum value of S is less than 0.5 so here we choose $s$ from $\{0.2, 0.4\}$. For step 3, $s$ has approximately uniform distribution across $[0,1]$, so to determine some distinct values here we choose S from {0.3, 0.8}. For step 4 as shown in figure 5d, $s$ gets higher value in [0 1] so we choose $s$ from $\{0.5, 0.9\}$. In this step blocks' size are $2 \times 2$ that cause to be encoded very well.

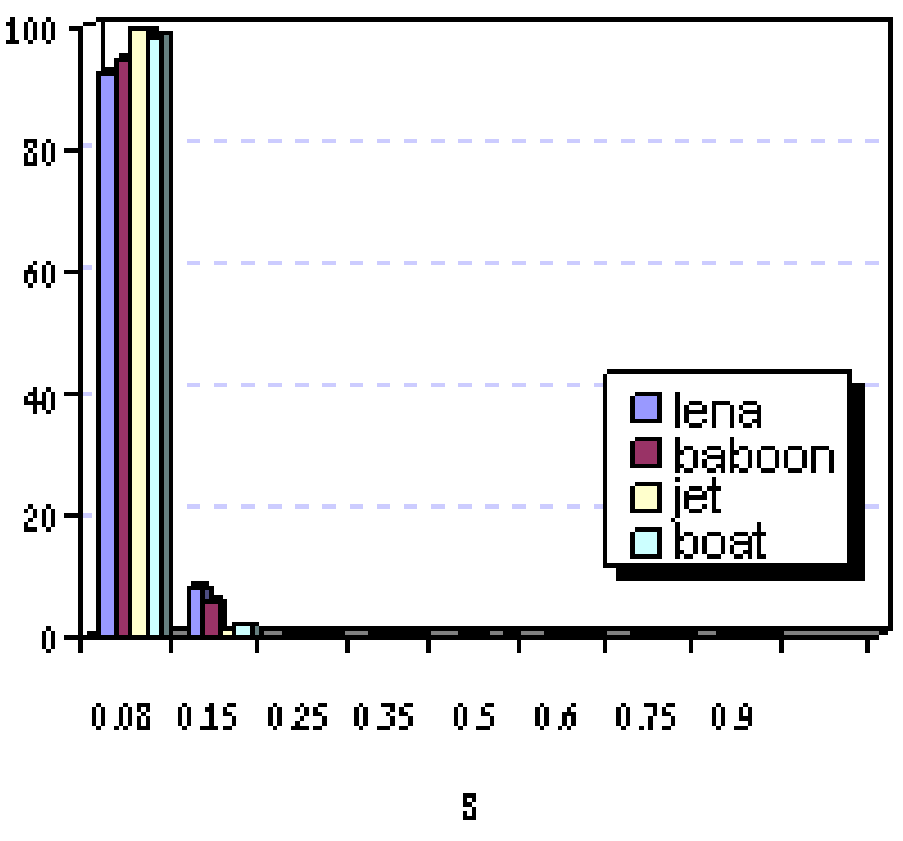

**(5a)**

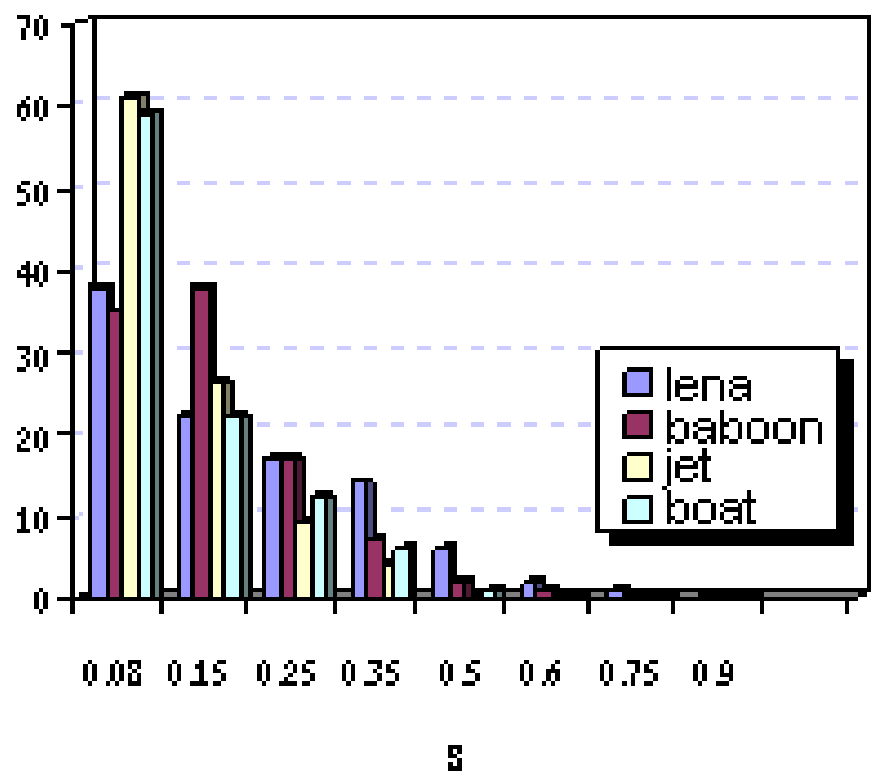

**(5b)**

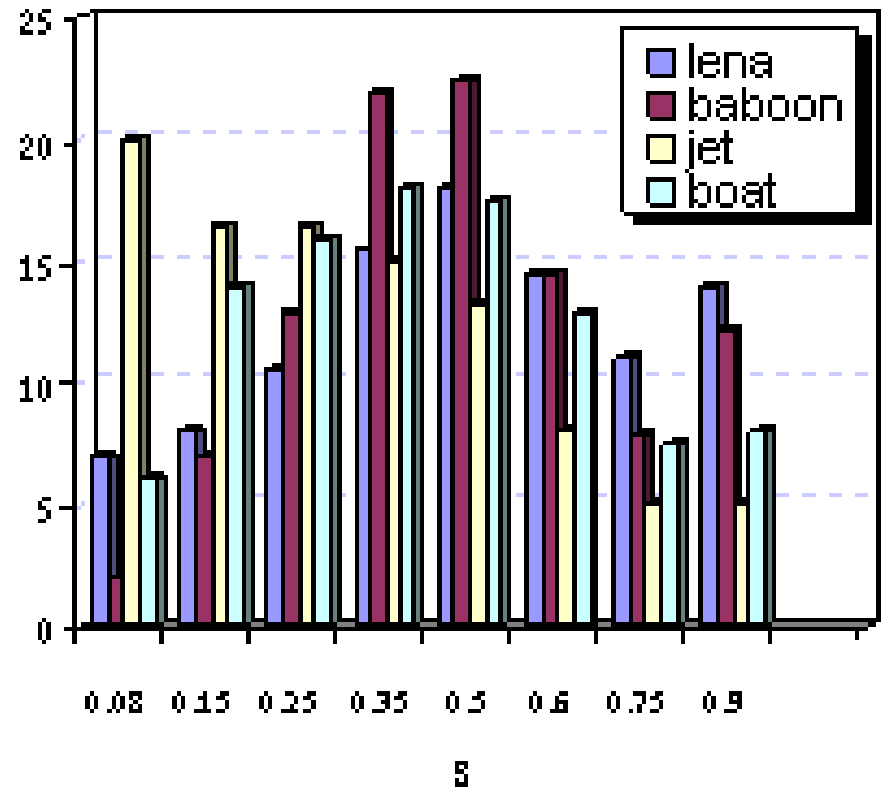

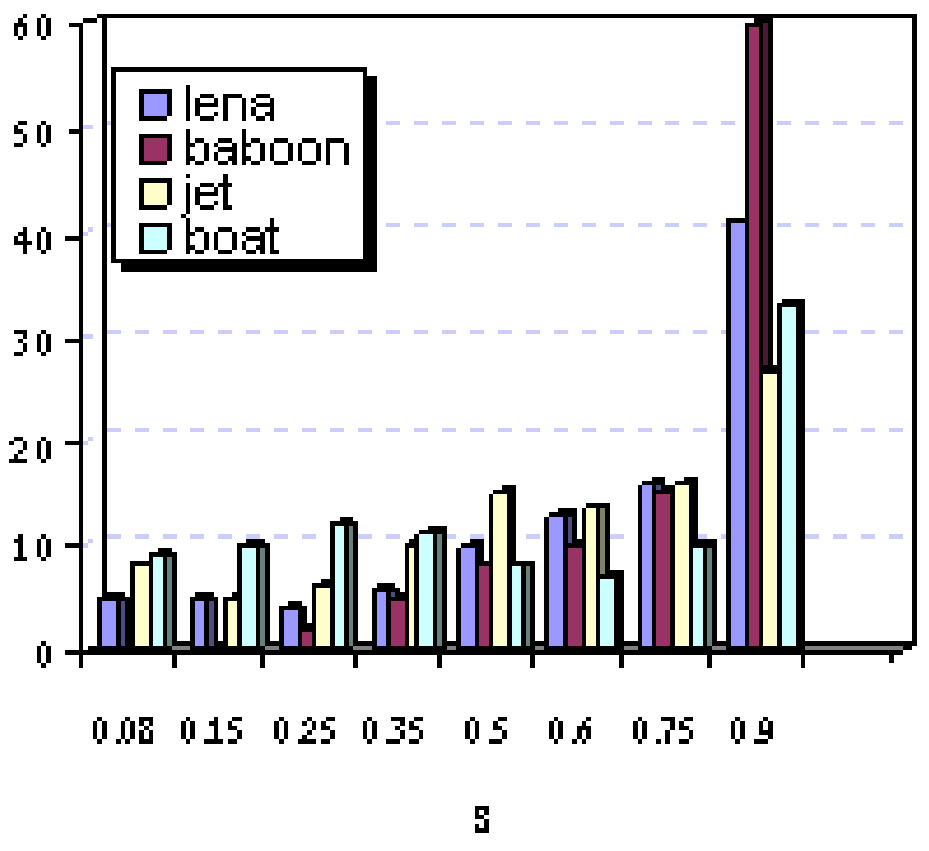

**(5d)**

**Figure 5 Histogram of $S$ at a) step1 b) step2 c)step 3 d)step 4**

We reduced the set of values of $s$ to a two- member set obtained above ({0.2,0.4}, {0.3,0.8},{0.5,0.9}) that leaves three cases for range blocks. The first case is where the selected value is the same value as gained from exhaustive search of $s$ here isn't any problem.

The second case is where the selected $s$ is not the best value, but the error is less than the threshold and the range blocks are coded approximately optimal here the encoding time is saved but the PSNR is somewhat damaged. In the third case, the selected value causes the encoding error is so large that the range blocks can not be coded here. So the range blocks are split and the encoding is done in next steps that leave better PSNR at the cost of small degradation of time and compression ratio.

### 3.2.1. The effect of pool size on the $s$

As mentioned above, we did our experiments with domain pool of size 256, 64 and 32. Figure 6 shows the histogram of selected $s$ with pool size as parameter for steps 1 and 2. It is obvious from figure 6 that for greater pool size the histograms are shifted left, on the other words for small pool size the $s$ will have smaller values. As shown in figure 6 the height of the histogram related to smaller pool size is greater than the other at small values of $s$ where at bigger pool size it isn't so. This point can be explained by this fact that when we make domain pool small indeed the highest entropy domain blocks will remain in domain pool and the difference between entropy of domain blocks and the range blocks will be high, and so $s$ has smaller Values to do matching. How much pool size is small then the histogram be of $s$ will be shifted left and thus restriction $s$ to one or two distinct values will create lesser error. Thus it seems that for small domain pool our algorithm works better than entropy based algorithm [12].

### 3.3 Data structure

In matching a domain and a range block, the mean value of range block is directly stored and we use only $S$ and the transformation. Here indeed we don't use equation (3) and this is a bit difference between our algorithm and the traditional forms. Figure 7 shows the associated data of a range block that is stored as compressed data.

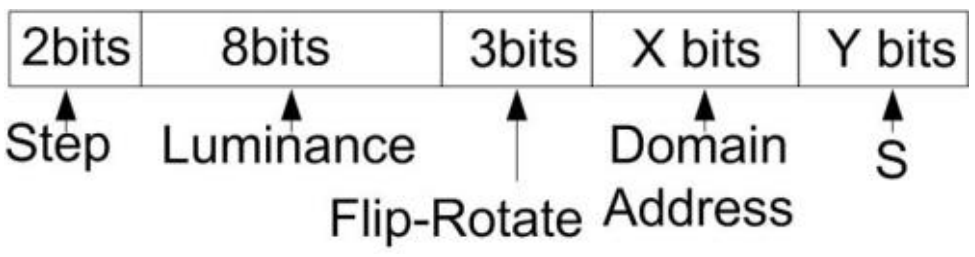

*Fig. 7 stored data format of a range block*

As shown above data of each range block consist of 5 fields; field 1 is the step number in which the block was coded and has 2 bit (the algorithm has four steps), the second field is the mean value of the range block, O, with 8 bits (0 to 255). Field 3 is 3 bit long and represents the isometric transformations (rotate and mirror). X bits for the address of associating domain block. X is given as follows,

$$X = \lceil \log_2^{DN} \rceil \qquad (8)$$

$\lceil x \rceil$ *is the smallest integer greater than x*

Where, $DN$ is the size of domain pool. For example if $DN = 64$ then 6 bits are used for the field. Here the coordination of domain blocks are kept in a list and the entries of the list are used as the address. It is clear that the addresses of range blocks need not be stored because of their regular order in the image and so in stored data. We store them left to right and up to down direction. The last field is $S$ with 1 bit length for steps 2, 3 and 4. (Due to two predefined value for $S$). The stored data frames don't contain this field at step 1, because here we only have one default value for $S$.

## 4. Experiments and results

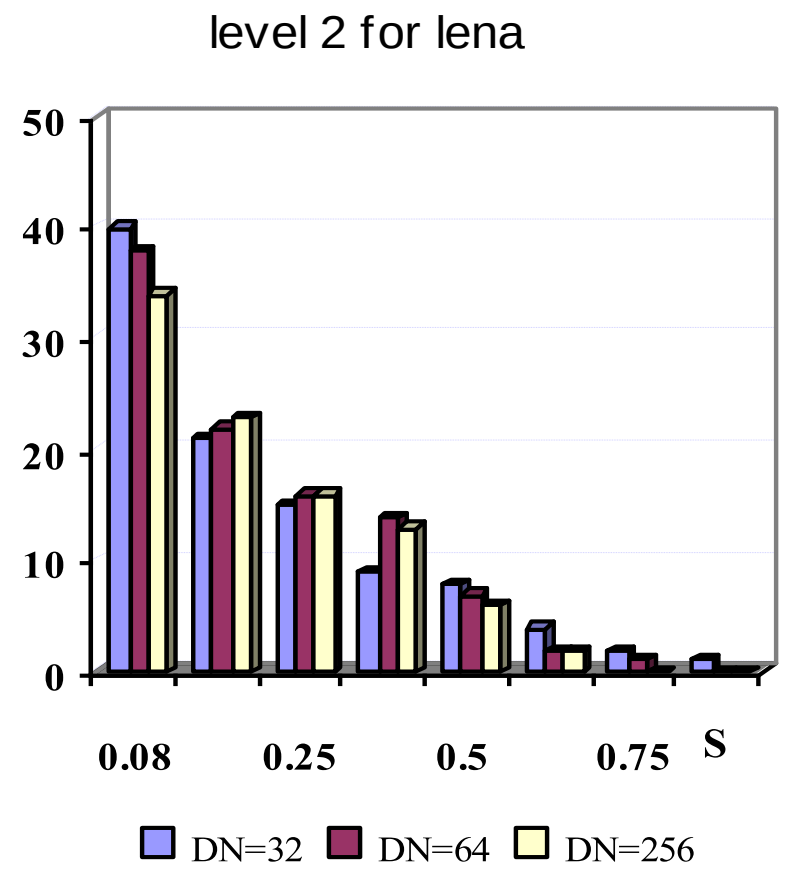

**(6b)**

***Figure 6, Histogram of $S$ with pool size as parameter for a) step 1 b)step 2***

We did different experiments to evaluate proposed algorithm and compare it with entropy based. These all were done in C++ on a Pentium 2 (450MHz) with 256 MB RAM. Comparison results are shown in figure 5a, b, c and d for different pool size and the Lena image. To have reasonable comparison the two algorithms are compared in fixed PSNR. Figure 8a,b show the compression ratio and encoding time for PSNR=33.71db, and Figure 8c,d show the compression ratio and encoding time for PSNR=35.07db.

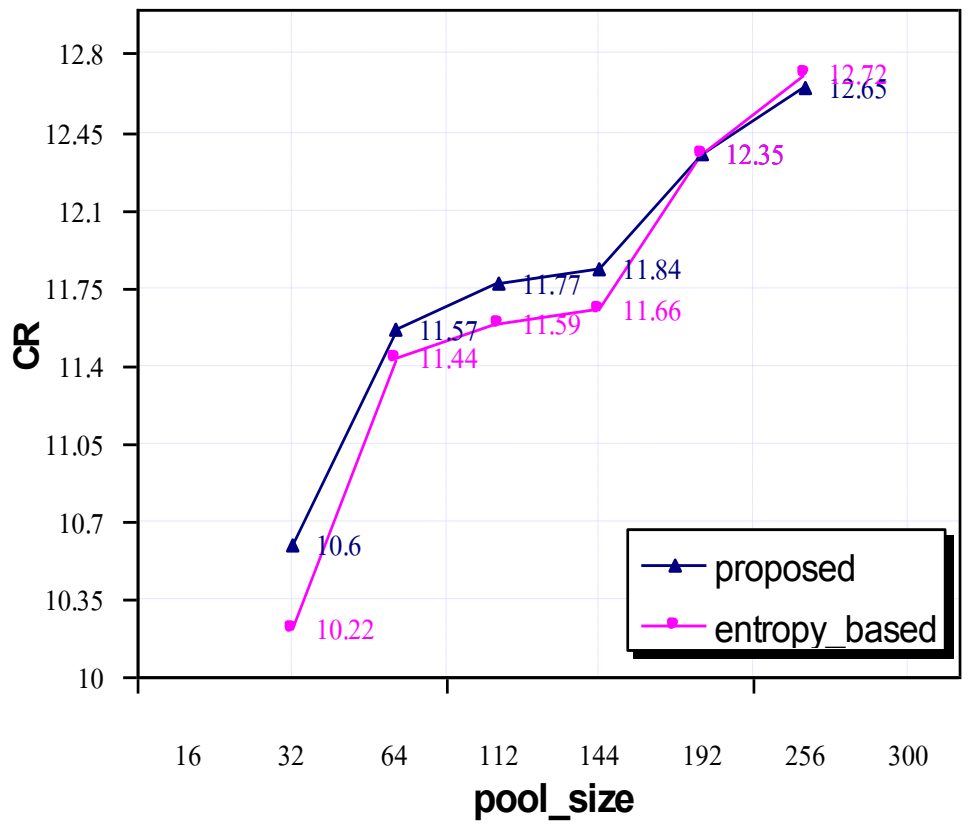

(8a)

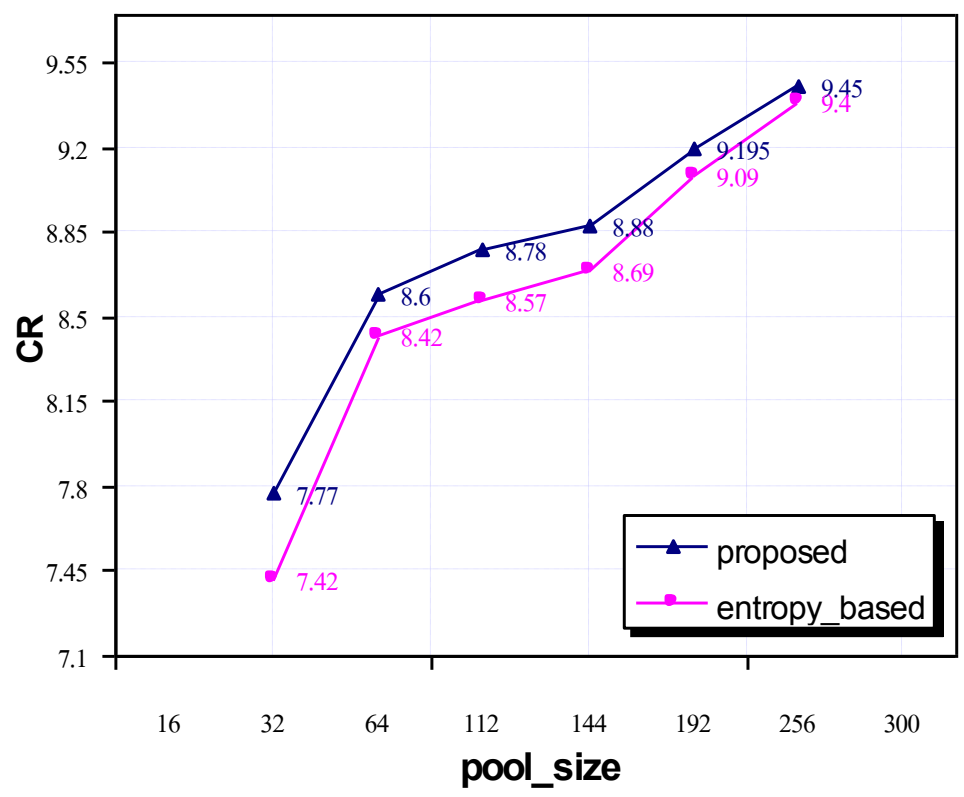

(8c)

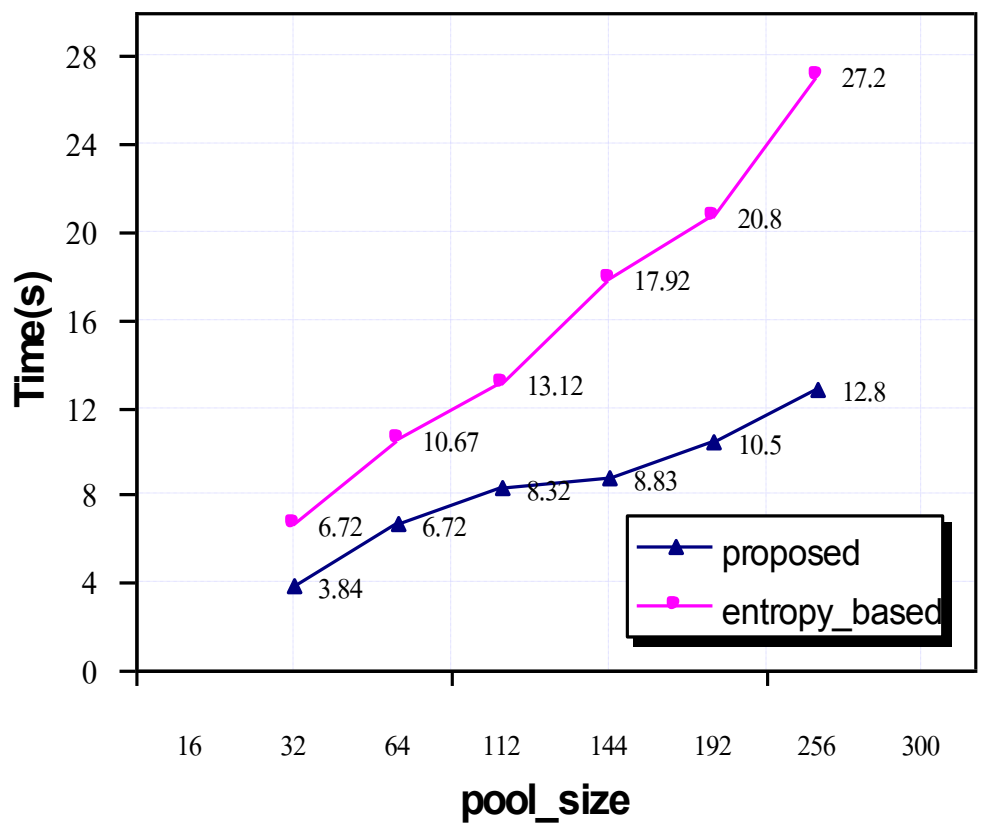

(8d)

Figure 8, Compression ratio and encoding time of proposed algorithm and entropy based versus pool size a, b) at fixed PSNR=33.71db, c, d) at fixed PSNR=35.07db

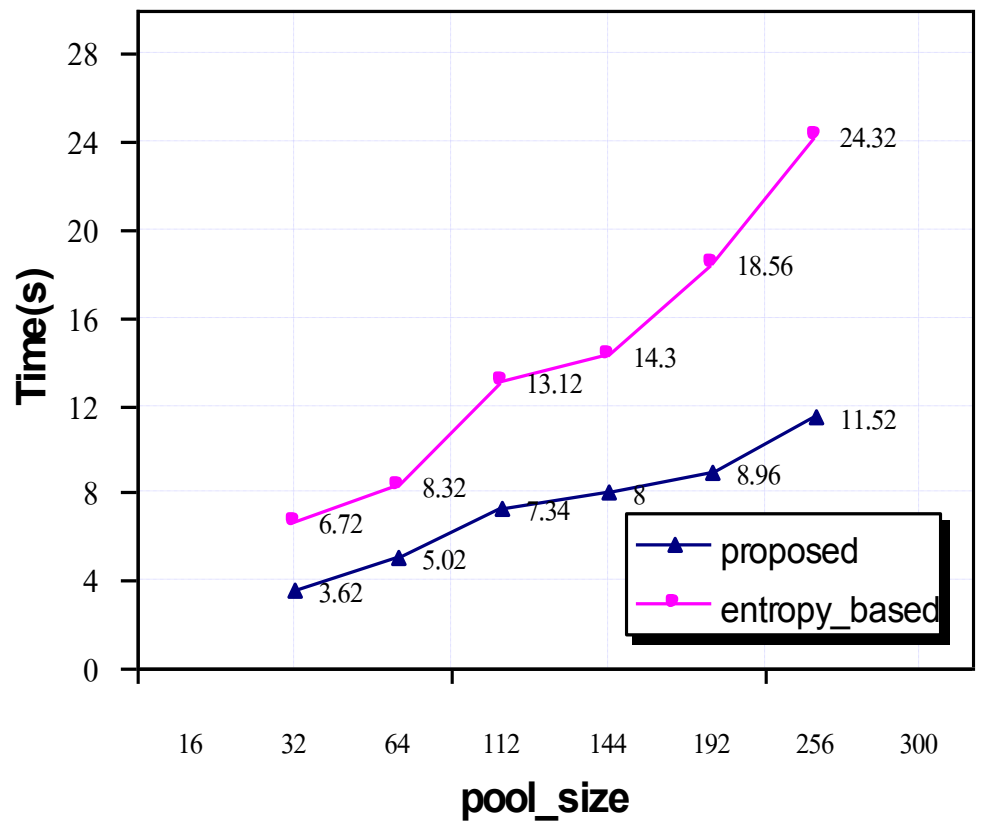

(8b)

In this figures compression ratio and encoding time are plotted versus pool size with the PSNR as parameter.

Figures 8a,b shows the results for PSNR=33.71db as a high quality of encoded image. As shown in figures 8a and b proposed algorithm works better when pool size is less than 192, comparing compression ratio, but the encoding time of proposed algorithm is better than the entropy based for all pool size. The time of proposed algorithm is close to entropy based at small pool size and by increasing pool size the difference becomes high. To explain this fact we can say that for each range block all domain block are investigated to find the best mapping. Suppose $t_{ent}, t_{prop}$ be the time needed to match a range block and a domain block in entropy based and proposed algorithm respectively. The total time to find best mapping that is done by evaluating all domain blocks will be proportional to pool size. If $T_{ent}, T_{prop}$ be the encoding time of the two algorithms we can simply write:

$$T_{ent} - T_{prop} \propto Pool_size(t_{ent} - t_{prop}) \qquad (9)$$

Where, $t_{ent} - t_{prop}$ is constant and positive.

For large domain pool, of size greater than 192, the entropy based algorithm might be better in view of compression ratio. Indeed when we increase the pool size the domain pool will include domain block of lower entropy (recall we choose blocks with the highest entropy and extending it means choosing lower entropy blocks) and here the histogram of $s$ based on former discussion will spread over $[0,1]$ at all four steps, so that restricting $s$ to predefined values will cause some blocks go to lower steps while they could be encoded in current step with $s$ different from selected one. As a flash result our algorithm is very suitable for fast applications. Tables 2 and 3 show also another comparison between the entropy based algorithm and the proposed algorithm for different pool sizes but this time for lower PSNR values.

Table 2 shows the results for entropy based algorithm. As shown and would be expected, increasing the pool size causes compression and PSNR both are increased and also the time of encoding is raised. Comparing the two algorithms in different experiments one can simply see that the proposed algorithm is better especially in encoding time which is less than 50% of another.

***Table 2: entropy based***

| pool size | Comp-ratio | Encod-time(S) | PSNR (db) |
|---|---|---|---|
| 32 | 10.12 | 5.89 | 33.75 |
| 32 | 10.98 | 5.28 | 33.36 |
| 64 | 11.8 | 8 | 33.55 |
| 64 | 10.96 | 9.6 | 33.90 |
| 144 | 12.17 | 16.32 | 33.57 |
| 144 | 11.3 | 17.28 | 33.75 |
| 256 | 13.17 | 26.24 | 33.70 |
| 256 | 11.97 | 28.16 | 34.80 |

***Table 3: proposed algorithm***

| pool size | Comp-ratio | Encod-time(S) | PSNR (db) |
|---|---|---|---|
| 32 | 10.87 | 4.32 | 33.60 |
| 32 | 11.07 | 4.25 | 33.52 |
| 64 | 11.47 | 4.8 | 33.78 |
| 64 | 11.7 | 5.12 | 33.9 |
| 144 | 12.18 | 8.38 | 33.63 |
| 144 | 12.52 | 8.32 | 33.49 |
| 256 | 12.67 | 13.12 | 33.83 |
| 256 | 13.21 | 13.12 | 33.61 |

***Table 4 the comparison results for F16***

| | Proposed | | | Entropy based | | |
|---|---|---|---|---|---|---|
| Pool size | Com. rat | Enc. Time (S) | PSNR | Com. rat | Enc. Time | PSNR |
| 256 | 11.25 | 13.76 | 33.41 | 11.50 | 85 | 33.41 |
| 64 | 9.63 | 5.44 | 33.87 | 9.47 | 25 | 33.97 |
| 32 | 9.66 | 4.16 | 33.65 | 9.5 | 21 | 33.64 |

***Table 5 the comparison results for Baboon***

| | Prop | | | Entropy | | |
|---|---|---|---|---|---|---|
| Pool size | Com. rat | Enc. Time (S) | PSNR | Com. rat | Enc. Time | PSNR |
| 256 | 5.36 | 26.24 | 26.33 | 5.35 | 47.04 | 26.34 |
| 64 | 4.98 | 10.88 | 26.40 | 4.84 | 18.56 | 26.09 |
| 32 | 4.62 | 7.68 | 26.07 | 4.61 | 12.16 | 26.07 |

***Table 6 the comparison results for Boat***

| | Prop | | | Entropy | | |
|---|---|---|---|---|---|---|
| Pool size | Com. Rat | Enc. Time (S) | PSNR | Com. rat | Enc. Time | PSNR |
| 256 | 11.35 | 12.48 | 31.24 | 11.36 | 25.92 | 31.24 |
| 64 | 10 | 6.08 | 31.30 | 10 | 9.44 | 31.30 |
| 32 | 9.8 | 4.16 | 31.30 | 9.65 | 6.4 | 31.30 |

To have good perception of proposed algorithm the results for three other familiar images are presented in tables 4, 5 and 6. At last for more insight on the efficiency of the algorithm, the PSNR of the algorithm is plotted versus compression ratio and is compared with the no search algorithm.

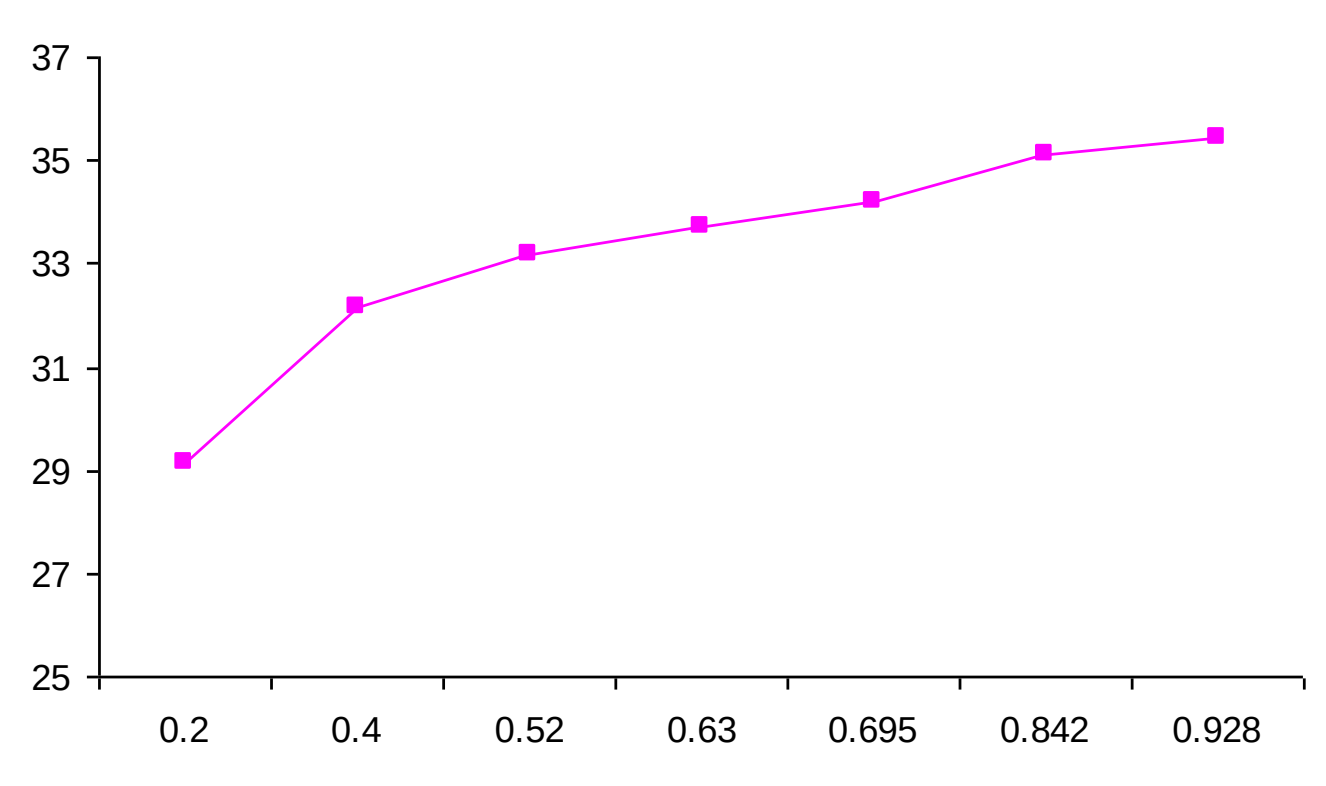

Figure 9. PSNR versus bpp for the proposed algorithm (applied on Lena)

## 5. Discussion about tradeoffs

As one can simply deduce there is a strong dependency between error threshold of steps and the compression ratio. If error thresholds have small values the algorithm for more ranges goes to lower steps (2, 3 and 4) this increases the amount of data stored, having more fidelity and being time consuming.

On the other hand proper values for error thresholds differ from one image to another. For example in baboon the blocks have very high variance and naturally error of mapping will be high so that choosing small values for thresholds causes the algorithm goes to lower steps approximately for almost all of the range blocks that leaves a very time consuming algorithm with very small compression ratio. In contrary to baboon, Lena's blocks have small variances and unlike baboon here error of mappings are very small, so that choosing high values for thresholds causes the reconstructed image have poor quality and ringing be evidence. On the other words regions of the Lena contain lower frequencies and small degradations are detected..

## 6. Conclusions and future works

In this paper we presented a new method for fractal image compression to reduce encoding time. Before anything we analyzed the effect of different parameters on different performance measures such as encoding time, compression ratio and PSNR. The most important point that decreased the encoding time was the way to not search the contrast scaling factor and use predefined values. Also our analysis showed that domain pool reduction has good performance for pool size less than 200 that leads to short times.

Experimental results show proposed method is better than previous method based entropy. In future we tend use this approach in frequency domain and compare with other hybrid method.

## Appendix A: Some samples of results

To have better understanding the result of the proposed algorithm here some image samples are presented:

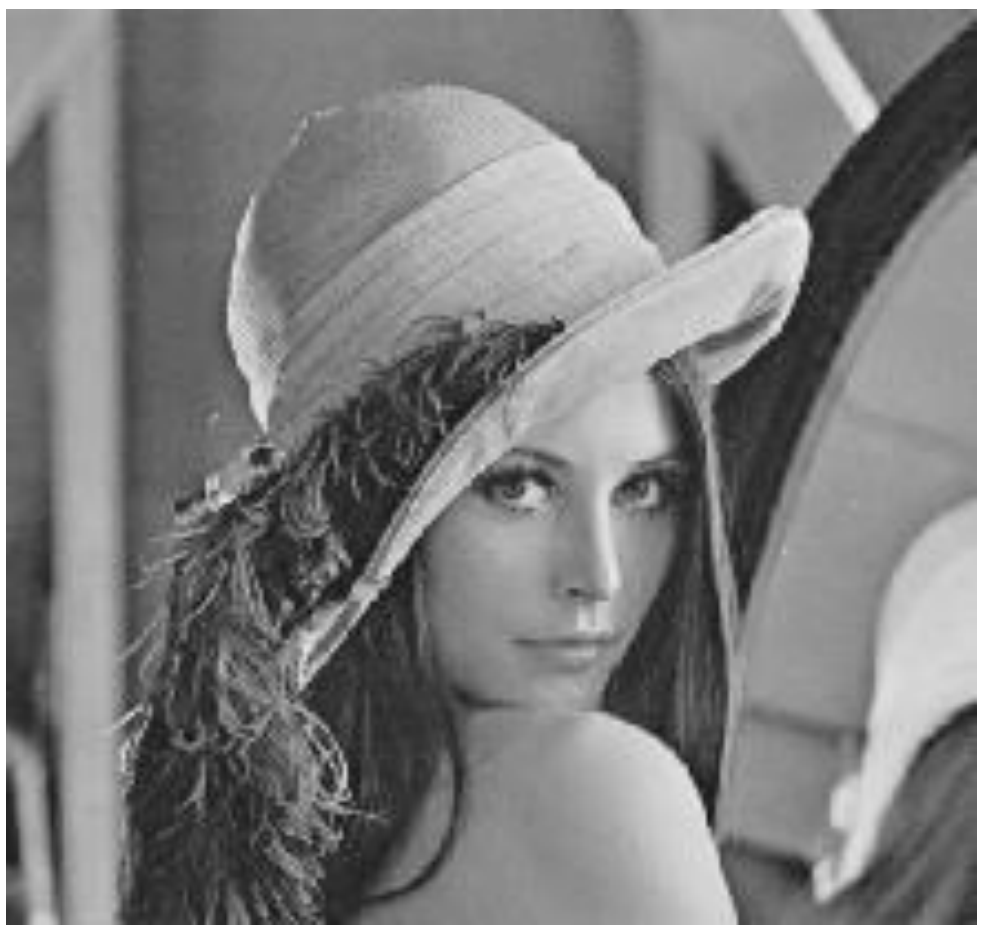

**Original Image**

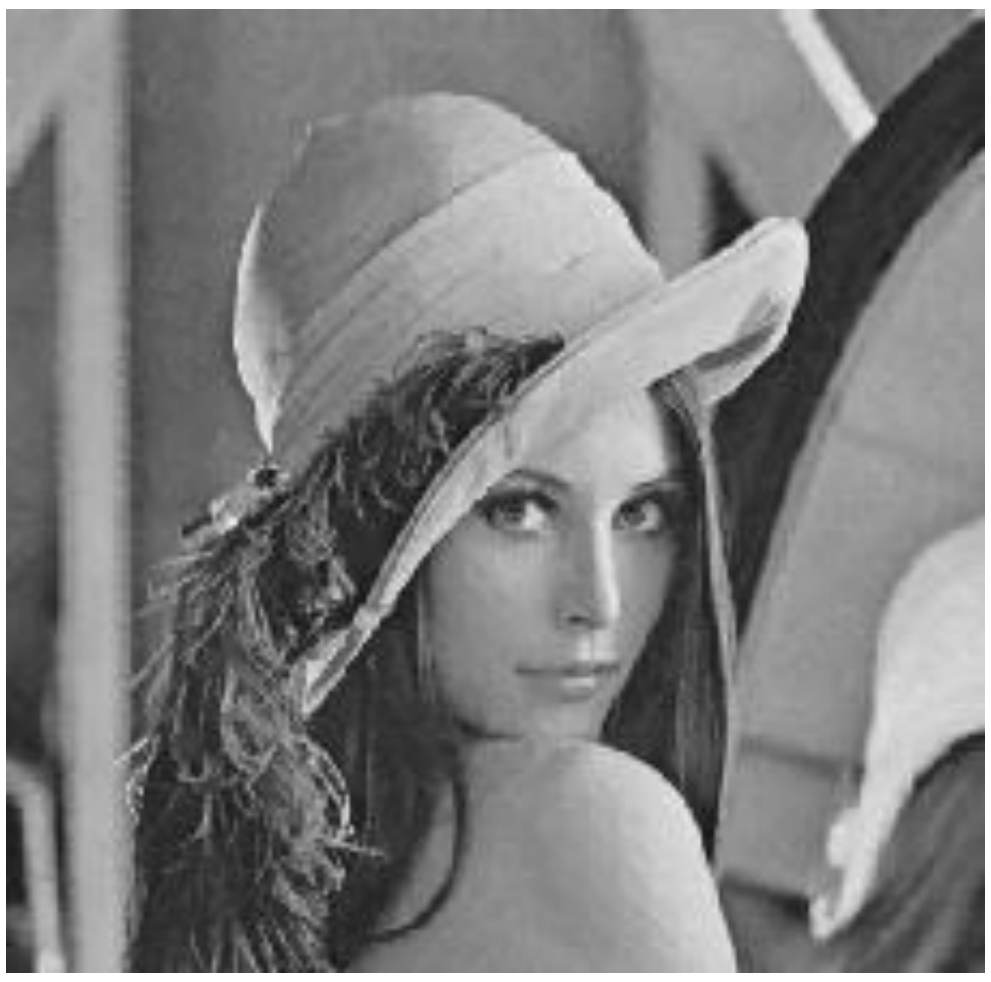

**Com.Rat=12.17 Time(8.2 s) PSNR=33.57db**

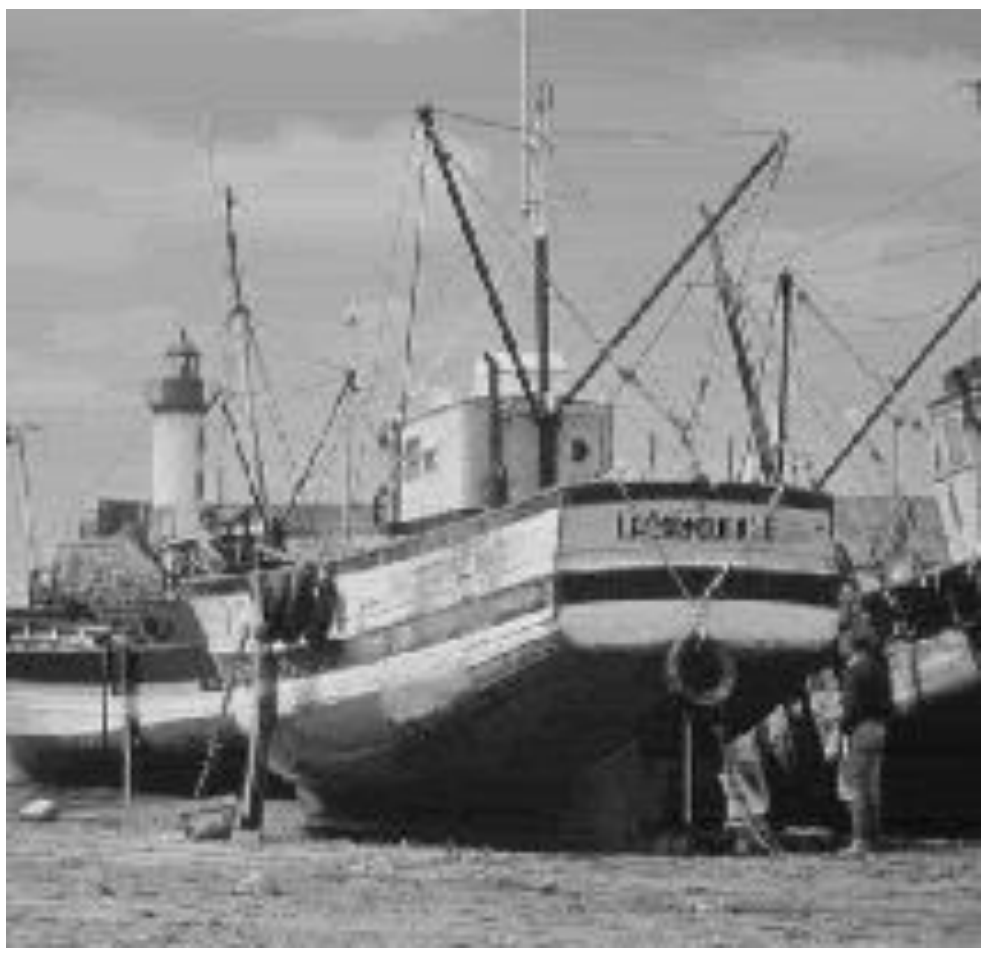

**Original Image**

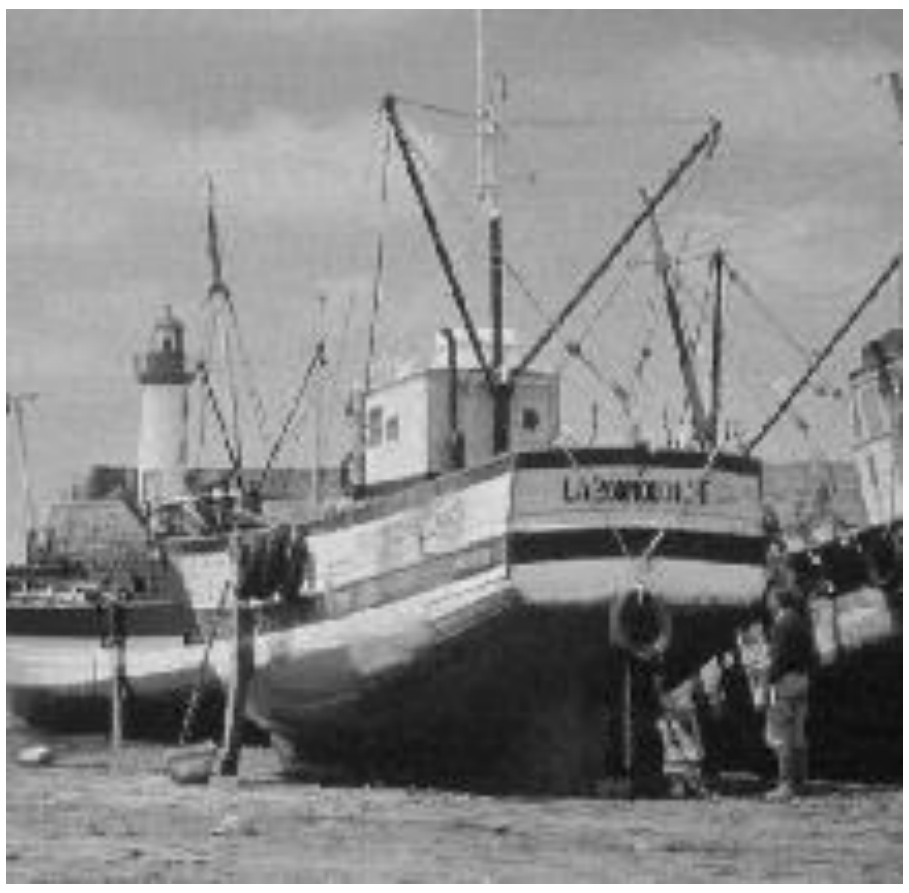

**Com.Rat=10 Time(6.08s) PSNR=31.30db**

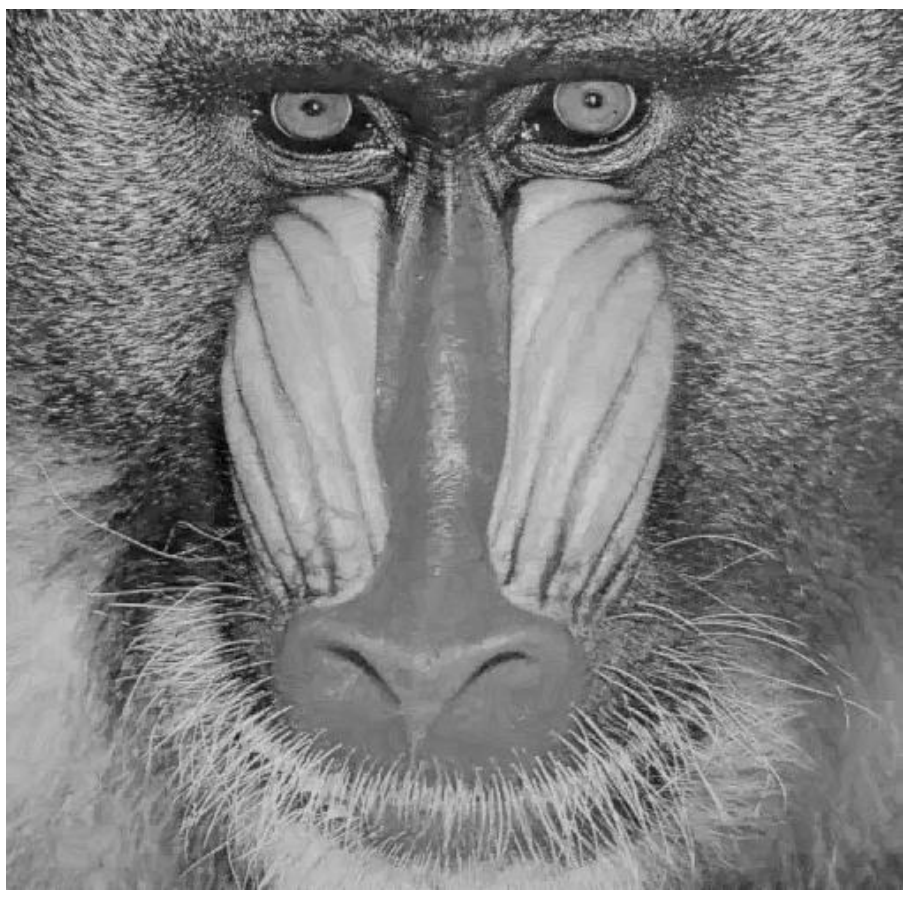

**original Image**

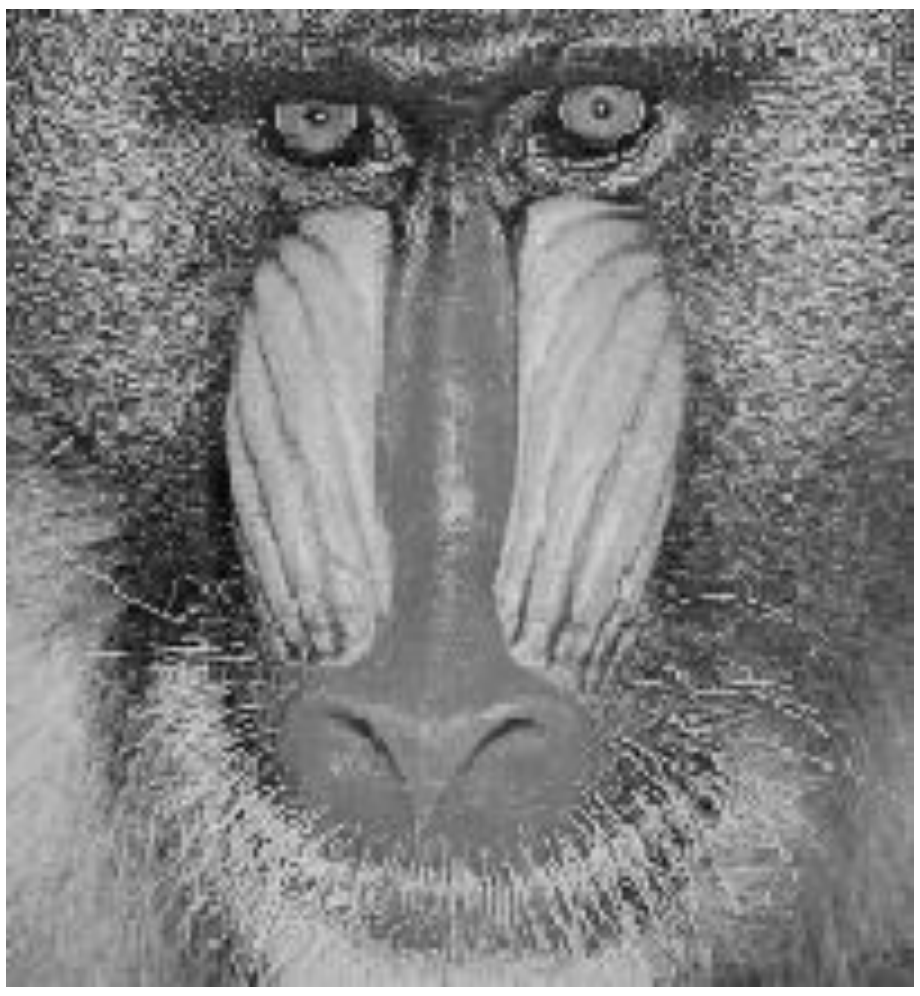

**PSNR=26.09db Com.Rat=4.8 Time(8.05 s)**

## References

[1] N.T.Thao, "A hybrid fractal–DCT coding scheme for image compression," in Proceeding ICIP-96 (IEEE International Conference on image processing), Lausanne, Switzerland, Sept. 1996, vol. 1, pp. 169-172.

[2] M. Kaplan and C.-C. J. Kuo, "Texture segmentation via haar fractal feature estimation," J. Vis. Commum. Image Represent, vol .6, no. 4, pp. 387-400, 1995.

[3] C.S. Tong, M. Pi, Analysis of a hybrid fractal Predictive coding compression scheme, signal Processing : Image Communication 18 (2003) 483-495.

[4] D. Saupe. Lean Domain Pools for Fractal Image Compression. Proceedings IS&T/SPIE 1996 Symposium on Electronic Imaging: Science & Technology Still Image Compression II, Vol. 2669, Jane 1996

[5] G. E. Qien, S. Lepsqy, and T. A. Ramstad, "An inner product space approach to image coding by contractive transformation s," in Proceeding ICASSp-91 (IEEE-International Conference on Acoustics Speech and Signal Processing) ‘ Toronto, Canada, May 1991, vol.4, pp. 2773-2776.

[6] T. Tan and H. Yan, "Object recognition using fractal neighbor distance :Eventual convergence and recognition rates , " in Proc. ICPR2000, Vol .2, 2000, pp. 785-788

[7] M. Barnsley and L. Hurd. Fractal Image Compression. On Image Processing: Mathematical Methods and Applications. pp. 183-210, Clarendon Press, Oxford, 1997.

[8] D. C. Popescu and H. Yan, "MR image compression using iterated function system" Magn. Res. Image.,vol. 11, pp. 727-732,1993.

[9] G. E.Ǿien and G. Narstad, "fractal compression of ECG signal,"in proc.NATO ASI fractal image encoding and analysis, Y,Fisher, Ed., Trondheim, Norway, July 1995.

[10] .W.Jacob, Y. fisher, R.D. Boss, Image compression: a study of the iterated contractive

image transform method, signal processing 29 (1992) 251-263.

[11] A.E.Jacquin, image coding based on a fractal theory of iterated contractive image transformation, IEEE Trans. Image Process .1 (1) (1992) 18-84.

[12] M.Hassaballah, M.M.Makky and Y.B. Mahdy, A Fast Fractal Image Compression Method Based entropy, Electronic Letters on computer Vision And Image Analysis 5(1):30-40,2005

[13] D.M.Monro, ”Class of fractal transforms,” Electronics Letters,vol.29,no.4,pp. 362-363,Feb. 1993.

[14] Y. Fisher. Fractal Image Compression: Theory and Applications. Springer-Verlag, New York, 1994.

[15] M. Salarian, H. Hassanpour. "A New Fast No Search Fractal Image Compression in DCT Domain.", Conf Proc international conference on machine vision, pp.62-66, 2007

[16] C.S. Tong, M. Wong, Adaptive approximate nearest neighbor search for fractal image compression, IEEE Trans. Image Process. 11 (6) (2002) 605–614.

[17] D. Saupe, Fractal image compression via nearest neighbor search, in: Conference on Proceedings of NATO ASI Fractal Image Encoding and Analysis, Trondheim, Norway, 1995.

[18] C.S. Tong, M. Pi, Fast fractal image encoding based on adaptive search, IEEE Trans. Image Process. 10 (9) (2001) 1269–1277.

[19] J. Li and C.-C. J. Kuo, “Image compression with a hybrid wavelet-fractal coder,” IEEE Trans. Image Process., vol. 8, no. 6, pp. 868–874, Jun. 1999.

[20] Y. Iano, A. Mendes and F. Silvestre, “A Fast and Efficient Hybrid Fractal-Wavelet Image Coder” IEEE TRANSACTIONS ON IMAGE PROCESSING, VOL. 15, NO. 1, JANUARY 2006, 97-105.